# Quad-cone-rotor: A Novel Tilt Quadrotor with Severe-fault-tolerant Ability

Zhe Shen, Yudong Ma, Takeshi Tsuchiya

The University of Tokyo   zesheng@g.ecc.u-tokyo.ac.jp

**Abstract**

Conventional quadrotors received great attention in trajectory design and fault-tolerant control in these years. The direction of each thrust is perpendicular to the body because of the geometrics in mechanical design. Comparing with the conventional quadrotor, a novel quadrotor named quad-tilt-rotor brings better freedom in manipulating the thrust vector. Quad-tilt-rotor augments the additional degrees of freedom in the thrust, providing the possibility of violating the normal direction of the thrust in the conventional quadrotor. This provides the ability of greater agility in control. This paper presents a novel design of a quad-tilt-rotor (quad-cone-rotor) whose thrust can be assigned along the edge of a cone shape. Besides the inheriting merits in agile from quad-tilt-rotor, the quad-cone-rotor is expected to take fault-tolerant control in severe dynamic failure (total loss in all thrusts). We simulate the control result in a UAV simulator in SIMULINK, MATLAB.

## I. INTRODUCTION

In a conventional quadrotor, the thrust is perpendicular to the body. The inputs are the angular velocity of each motor which generates the thrust to control attitude and the altitude. We should note that the number of the input (four) is less than the degrees of freedom (6). Thus, following an arbitrary trajectory with the specific requirement in attitude can be impossible for this under-actuated system [1]. A common method to deal with this under-actuated problem is to find the relationship between the actuated variables and the variables not actuated directly. [2-5] approximates this relationship using linearization and infinitesimal equivalence. Thus, the degree of freedom is reduced to four, which is equal to the number of the inputs.

Although the conventional quadrotor can be stabilized to track the desired trajectory [5-8], the inevitable attitude change in translational movement can influence the onboard camera. Thus, the camera-based landing task can be a challenge. With these considerations, novel quadrotors named quad-tilt-rotor or tilted rotor UAV are developed [9-12]. Compared with the conventional quadrotor, the quad-tilt-rotor can adjust the direction of its thrust; several extra servo motors make it possible.

There are two typical designs of these quad-tilt-rotors. One is the quad-tilt-rotor with eight inputs (four for the magnitude of the thrust; four for the direction of the thrust) [9]. The four servo motors provide the possibility of changing the direction of each thrust by rotating them along each arm. The other quad-tilt-rotor owns twelve inputs [10]. The function of the eight motors among its twelve motors is identical to the function of the eight inputs in the previous quad-tilt-rotor. The extra four motors are used to further adjust the direction of thrusts, which provides more freedom in selection of the potential direction of the thrust.

The promising fact is that the number of the inputs (eight or twelve) is more than the quantity of the variables (six). Indeed, the quad-tilt-rotor becomes an over-actuated system with input redundancy, which brings the possibility of tracking the desired trajectory and the designed attitude at the same time [13]. However, the complexity introduced by the extra inputs burdens the development of the proper controller to manipulate these novel quad-tilt-rotors. A typical method in controller design is to decrease the number of degrees of freedom by restricting the allocation of the servo motor on purpose [2,3,14].

Another notable advantage of quad-tilt-rotors is its ability in fault-tolerant control [4,11,12,14-16]. The redundancy gifted by the tilt mechanism entails the different fault-tolerant control results with the conventional UAV. The common faults studied are partially or complete loss in one or several thrusts [15-18]. Note that the total loss in all thrusts is beyond the scope of these studies; these quad-tilt-rotors are determined to fall down without the possibility of getting controlled.

This research proposes a novel design of a quad-tilt-rotor whose degree of freedom is eight. The direction of the thrust of each rotor can be selected along the surface of a cone shape. Thus, we name it quad-cone-rotor. The dynamics are deduced based on Newtown-Euler equations. A simulator is subsequently modeled based on these dynamics in SIMULINK, MATLAB. Symmetrical open-loop input test is conducted for both cases where the thrusts are healthy and where the driven-thrust motors are in stuck (total loss in all thrusts).

This paper is structured as follows: Section II illustrates the design and the dynamics of the quad-cone-rotor. The model of the thrusts and drag moments is explained in Section III. The symmetrical open-loop hovering test and the fault-tolerant open-loop hovering test are detailed in Section IV and V, respectively. Section VI presents the trade-off analysis in designing a cone-angle for a quad-cone-rotor. At last, the conclusions and discussions are made in Section VII.

## II. Quad-cone-rotor Dynamic Model

This section is to provide the kinematics design and the dynamical model of the novel quad-cone-rotor. Before illustrating the kinematics, we start with the introduction to the designation at the end of each arm.

### *Direction of The Thrust*

To better introduce the design of our quad-cone-rotor, we refer to the quad-tilt-rotor designed in [9]. As plotted in Figure 1 in [9], they proposed the structure where the direction of the thrust can be selected perpendicular to each arm.

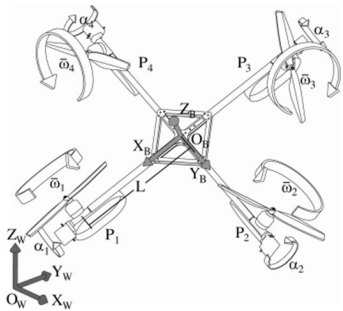

Fig. 1. The design of a quad-tilt-rotor in [9]

We plotted the direction of the potential thrust in blue in Figure 2. These potential thrusts constitute a sector, assuming that the thrusts are equal. Note that this sector is in a plane perpendicular to the arm.

Apart from the blue sector, there are three red cones in Figure 2. These potential cones are the designed potential thrusts in our quadrotor. The designed thrust can also change its direction, in a unique pattern (along the surface of a cone), however. We decide one cone in Figure 2 as the 'orbit' for our thrust by deciding on a proper cone-angle. The relevant consideration in picking this cone-angle is detailed in Section V.

We name this kind of quad-tilt-rotor quad-cone-rotor because of the cone shape of its potential direction of the thrust.

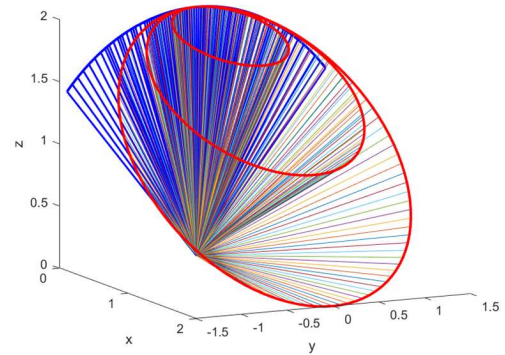

Fig. 2. The blue sector is the potential thrusts in [9]. The red cones are the potential thrusts in our quad-cone-rotor.

The thorough mechanical fulfillment of this quad-cone-rotor is beyond the scope of this study. However, we propose the draft of one possible design (Figure 3) to explain the kinematics of our quad-cone-rotor. Figure 3 is the structure at the end of each arm. There are two motors, the 'cone motor' and the 'rotor motor'.

The cone motor is fixed at bottom. It can work as a servomotor adjusting the direction of the thrust. Specially, the link fixed on the top of the cone motor is not parallel to the axis of the cone motor. Instead, there is an angle $\phi$ biased from the axis of the cone motor. This design determines the zone (cone shape) where the potential directions of the thrust can be. Section V explain the selection of this cone-angle, $\phi$.

The other motor which directly drives the rotor is called 'rotor motor'. It mainly determines the magnitude of the thrust if the coupling effect of the cone motor is not considered.

The length of the link between the cone motor and the rotor motor is $d$. The orientation of this link is carefully designed; it precisely points upward when the cone motor is at the left-most position.

It is worth mentioning that the cone motor has two modes. One is working as a servomotor; the other mode is the speed motor. While the rotor motor has the speed motor mode only.

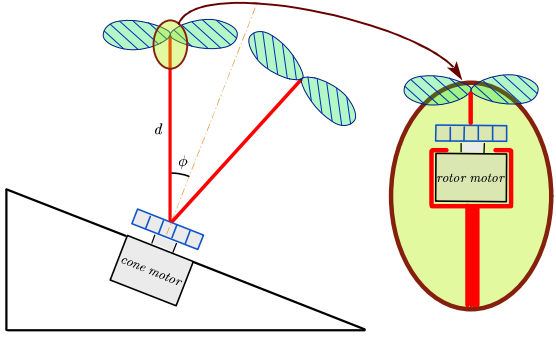

Fig. 3. The structure at the end of each arm.

At the end of each arm fixes this structure. So that the direction of each thrust can be along a cone. We assume that the angle of the cone for each thrust is identical ($\phi$).

**Kinematics**

Several coordinate frames are demonstrated in Figure 4. For simplicity and without losing generality, we only plot typical frames for one arm. The frames on the rest arms are similar.

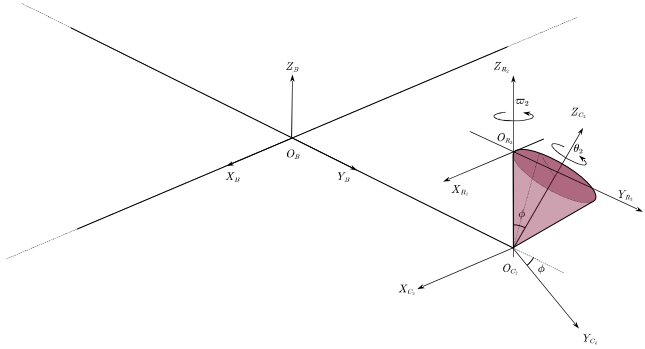

Fig. 4. The frames in the quad-cone-rotor.

Let $\mathcal{F}_B:\{O_B; X_B, Y_B, Z_B\}$ be the body-fixed frame. It is fixed at the geometrical center of the quad-cone-rotor. The $X_B$ is along the first arm; the $Y_B$ is along the second arm. The number of the arm is assigned in CCW (counter clockwise).

$\mathcal{F}_{C_2}:\{O_{C_2}; X_{C_2}, Y_{C_2}, Z_{C_2}\}$ is the frame fixed on the second cone motor. It is transformed from the $\mathcal{F}_B$ by rotating along $X_B$ by angle $-\phi$. The negative sign represents the direction in rotation is CW (clockwise).

The rotor frame $\mathcal{F}_{R_2}:\{O_{R_2}; X_{R_2}, Y_{R_2}, Z_{R_2}\}$ is fixed on the second rotor motor. Since $\mathcal{F}_{R_2}$ is not fixed on the propeller, it does not rotate with the propeller. However, it will change the position with respect to $\mathcal{F}_B$ if the cone motor changes the angle.

We assume the angle adjusted by the cone motor (fixed with $\mathcal{F}_{C_2}$) is $\theta_2$. Thus, the angular velocity of the cone motor is $\dot{\theta}_2$. Both directions of $\theta_2$ and $\dot{\theta}_2$ are CCW observed from the top of $Z_{C_2}$. In addition, the angular velocity of the rotor motor (fixed with $\mathcal{F}_{R_2}$) is assumed to be $\varpi_2$. The direction of $\varpi_2$ is CCW observed from the top of $Z_{R_2}$.

Similarly, the relevant frames, $(\mathcal{F}_{C_1}, \mathcal{F}_{R_1})$, $(\mathcal{F}_{C_3}, \mathcal{F}_{R_3})$, and $(\mathcal{F}_{C_4}, \mathcal{F}_{R_4})$, are defined at the end of the first, third, and fourth arm, respectively. The corresponding angles and angular velocities $(\theta_1, \varpi_1)$, $(\theta_3, \varpi_3)$, $(\theta_4, \varpi_4)$, are also defined. Note that we define the positive directions of angular velocities for both $(\theta_i, \varpi_i)$ in proposition 1.

*Proposition 1:* (positive directions of the angles and the angular velocities) The positive direction of the angular velocities, $(\theta_i, \varpi_i)$ at the end of the arm $i$ ($i=1, 2, 3, 4$), are defined in (1):

$$\begin{cases} \text{The positive directions of } Z_{C_i} \text{ and } Z_{R_i}, & i=2, 4 \\ \text{The negative directions of } Z_{C_i} \text{ and } Z_{R_i}, & i=1, 3 \end{cases} \quad (1)$$

For instance, the positive directions of $(\theta_2, \varpi_2)$ are along the positive direction of $Z_{C_2}$ and $Z_{R_2}$, respectively. While the positive direction of $(\theta_1, \varpi_1)$ are defined as the negative direction of $Z_{C_1}$ and $Z_{R_1}$, respectively.

We assume that $\varpi_i > 0$ generates the positive thrust (the positive direction of $Z_{R_i}$) by the definition. It means that CW rotation along $Z_{R_1}$ or $Z_{R_3}$ generates the positive thrusts. While the CCW rotation along $Z_{R_2}$ or $Z_{R_4}$ generates the positive thrusts. It is the consequence of the mechanical structure of the propellers.

We also restrict that

$$\varpi_i \geq 0, \quad i=1, 2, 3, 4 \quad (2)$$

We do not expect the negative angular velocity of the rotor motor. However, the negative angular velocities of the cone motors ($\dot{\theta}_i$) are allowed; the direction of each thrust can be adjusted by rotating the cone motor CCW or CW.

In the following text, we use $R_X$, $R_Y$, $R_Z$ to represent the rotation matrix along $X-axis$, $Y-axis$, $Z-axis$, respectively. In addition, we use $sA$ and $cA$ to represent $\sin(A)$ and $\cos(A)$, respectively.

The relevant matrix is demonstrated in Equation (3)-(5).

$$R_X(A) = \begin{bmatrix} 1 & 0 & 0 \\ 0 & cA & -sA \\ 0 & sA & cA \end{bmatrix} \quad (3)$$

$$R_Y(A) = \begin{bmatrix} cA & 0 & sA \\ 0 & 1 & 0 \\ -sA & 0 & cA \end{bmatrix} \quad (4)$$

$$R_Z(A) = \begin{bmatrix} cA & -sA & 0 \\ sA & cA & 0 \\ 0 & 0 & 1 \end{bmatrix} \tag{5}$$

The transformations between each frame are illustrated in (6)

$$\mathcal{F}_B \xrightarrow{R_Z\left(-\pi + \frac{\pi}{2} \cdot i\right) \cdot R_X(-\phi)} \mathcal{F}_{C_i} \xrightarrow{R_Z((-1)^i \cdot \theta_i) \cdot R_X(\phi)} \mathcal{F}_{R_i}, \quad i = 1, 2, 3, 4 \tag{6}$$

The notation, $^Y A_B^X$, represents the physical quantity, $A_B$, with respect to the frame, $\mathcal{F}_X$, expressed in the frame, $\mathcal{F}_Y$. When $\mathcal{F}_X$ is the inertial frame, $X$ is omitted (e.g., $^Y A_B$). When $\mathcal{F}_Y$ is the frame fixed on the body of $B$, $Y$ is omitted (e.g., $A_B^X$).

We use $^{R_i}\omega_{P_i}$ to represent the angular velocity of the propeller $i$ expressed in the rotor frame, $\mathcal{F}_{R_i}$. We use $\omega_{C_i}$ to represent the angular velocity of the cone motor $i$ expressed in its own frame (cone frame, $\mathcal{F}_{C_i}$). We use $\omega_B$ to represent the angular velocity of the body expressed in its own frame (body-fixed frame, $\mathcal{F}_B$). All of these angular velocities are with respect to the inertial frame.

We have the following relationship for $^{R_i}\omega_{P_i}$ in Equation (7).

$$^{R_i}\omega_{P_i} = {^{R_i}}w_B + {^{R_i}}\omega_{C_i}^B + {^{R_i}}\omega_{P_i}^{C_i} \tag{7}$$

Considering the relationship in (6), Equation (7) can be rewritten in Equation (8).

$$^{R_i}\omega_{P_i} = \left[R_Z\left(-\pi + \frac{\pi}{2} \cdot i\right) \cdot R_X(-\phi) \cdot R_Z((-1)^i \cdot \theta_i) \cdot R_X(\phi)\right]^T \cdot w_B$$
$$+ \left[R_Z((-1)^i \cdot \theta_i) \cdot R_X(\phi)\right]^T \cdot \omega_{C_i}^B + \begin{bmatrix} 0 \\ 0 \\ (-1)^i \cdot \varpi_i \end{bmatrix} \tag{8}$$

$w_B$ and $\omega_{C_i}^B$ are illustrated in Equation (9)-(10).

$$w_B = \begin{bmatrix} p \\ q \\ r \end{bmatrix} \tag{9}$$

$$\omega_{C_i}^B = \begin{bmatrix} 0 \\ 0 \\ (-1)^i \cdot \dot{\theta}_i \end{bmatrix} \tag{10}$$

Similarly, we have the relationship for $\omega_{C_i}$ in Equation (11).

$$\omega_{C_i} = \left[R_Z\left(-\pi + \frac{\pi}{2} \cdot i\right) \cdot R_X(-\phi)\right]^T \cdot \omega_B + \omega_{C_i}^B \tag{11}$$

***Dynamics***

The method in deducing the dynamic equation for our quad-cone-rotor in this study is Newtown-Euler method. The total movement of the vehicle comprises rotation and translation.

1) Rotation

Several studies (e.g., [2,10]) regard their targeted quad-tilt-rotor as an entirety without considering the effects brought by the rotation from the tilt rotor, ignoring the transient dynamics during the directional change in thrust. While [9] takes this effect into serious consideration by analyzing the dynamics in each frame.

We analyze the dynamics in each frame to describe the motion of our quad-cone-rotor in a detail way in this study. The relationship between each part is sketched in Figure 5. The drag (the cause for drag moment) and the thrust affect the propeller directly. These effects are described in the rotor frame $\mathcal{F}_{R_i}$.

The $rotor-motor_i$ links the rotor frame $\mathcal{F}_{R_i}$ and the cone motor frame $\mathcal{F}_{C_i}$, conveying the dynamical effect of the drag and the thrust. After that, the $cone-motor_i$ links the body-fixed frame $\mathcal{F}_B$ and the $\mathcal{F}_{C_i}$, further conveying the dynamical effect.

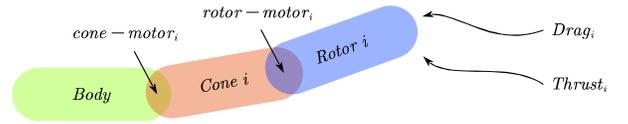

Fig. 5. The relationship between each part.

Several reasonable assumptions and approximations are made to simplify our modeling.

*Proposition 2:* (direction of the thrust) The direction of the thrust is perpendicular to the propeller and is affecting at the mid of the propeller. Thus, the thrust is in the positive direction of $Z_{R_i}$.

*Remark 1:* With proposition 2, the thrust contributes no moment to the $Rotor_i$

*Proposition 3:* (direction of the drag moment) The direction of the drag moment is perpendicular to the propeller and is affecting at the mid of the propeller. Furthermore, the thrust is in the negative direction of $Z_{R_i}$ (along $-Z_{R_i}$). Thus, it tends to decrease the angular velocity of the relevant rotor motor.

*Proposition 4:* (mass of the body) The mass of the body is much larger than the mass of the cone motor or the rotor.

Remark 2: With proposition 4, the thrust can be regarded as the force exerting directly on the body.

Remark 3: Based on Proposition 2-4 and Remarks 1-2, the thrust can is affecting on the body directly. Thus, we have the equivalent relationship for each part in Figure (6).

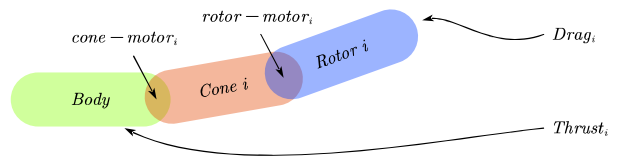

Fig. 6. The equivalent relationship between each part.

Based on this, we write the dynamic equation for each of these three separated parts based on Figure 5.

For $Rotor\ i$, we have Equation (12).

$$^{R_i}\tau_{P_i} = {}^{R_i}I_{P_i} \cdot {}^{R_i}\dot{w}_{P_i} + {}^{R_i}w_{P_i} \times ({}^{R_i}I_{P_i} \cdot {}^{R_i}w_{P_i}) - {}^{R_i}\tau_{ext_i} \quad (12)$$

$^{R_i}\tau_{ext_i}$ is the drag moment from the air. We will discuss it in detail later in Section III.

$^{R_i}\tau_{P_i}$ is the torque contributed by $Cone\ i$.

For $Cone\ i$, we have Equation (13).

$$\tau_{C_i} = I_{C_i} \cdot \dot{\omega}_{C_i} + \omega_{C_i} \times (I_{C_i} \cdot \omega_{C_i}) + R_Z\left((-1)^i \cdot \theta_i\right) \cdot R_X(\phi) \cdot {}^{R_i}\tau_{P_i} \quad (13)$$

$\tau_{C_i}$ is the torque contributed by $Body$.

For $Body$, we have Equation (14).

$$\tau_B = I_B \cdot \dot{\omega}_B + \omega_B \times (I_B \cdot \omega_B) + \sum_{i=1}^{4} R_Z\left(-\pi + \frac{\pi}{2} \cdot i\right) \cdot R_X(-\phi) \cdot \tau_{C_i} \quad (14)$$

$\tau_B$ is the torque provided by the thrusts. It is detailed together with $^{R_i}\tau_{ext_i}$ later in Section III.

2) Translation

The translational motion of the quad-cone-rotor is straightforward. It is described in Equation (15).

$$R_{W \to B} \cdot \sum_{i=1}^{4} {}^{B}T_i + \begin{bmatrix} 0 \\ 0 \\ -m \cdot g \end{bmatrix} = m \cdot {}^{W}\ddot{p} \quad (16)$$

$R_{W \to B}$ in Equation (16) is a rotation matrix transforming the thrust expressed in the body-fixed frame into the thrust expressed in the inertial frame. It is detailed in Equation (17).

$$R_{W \to B} = \begin{bmatrix} c\theta \cdot c\psi & s\varphi \cdot s\theta \cdot c\psi - c\varphi \cdot s\psi & c\varphi \cdot s\theta \cdot c\psi + s\varphi \cdot s\psi \\ c\theta \cdot s\psi & s\varphi \cdot s\theta \cdot s\psi + c\varphi \cdot c\psi & c\varphi \cdot s\theta \cdot s\psi - s\varphi \cdot c\psi \\ -s\theta & s\varphi \cdot c\theta & c\varphi \cdot c\theta \end{bmatrix} \quad (17)$$

Where $\varphi, \theta, \psi$ are the roll, pitch, yaw, respectively.

The relationship [19] between roll, pitch, yaw and the angular velocities is given in Equation (18).

$$\begin{bmatrix} \dot{\varphi} \\ \dot{\theta} \\ \dot{\psi} \end{bmatrix} = \begin{bmatrix} 1 & \sin(\varphi) \cdot \tan(\theta) & \cos(\varphi) \cdot \tan(\theta) \\ 0 & \cos(\varphi) & -\sin(\varphi) \\ 0 & \sin(\varphi) \cdot \sec(\theta) & \cos(\varphi) \cdot \sec(\theta) \end{bmatrix} \cdot \begin{bmatrix} p \\ q \\ r \end{bmatrix} \quad (18)$$

***Simulator***

Based on the aforementioned relationships in the kinematics and the dynamics, we build the simulator for our quad-cone-rotor in SIMULINK, MATLAB, environment (Figure 7).

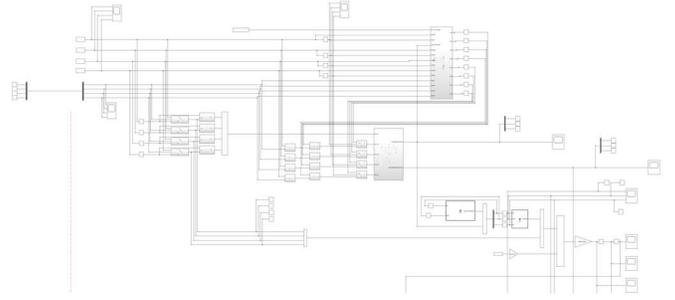

Fig. 7. The simulator of the quad-cone-rotor.

As can be seen from Figure 7, there are eight inputs in total. Four of them are the angular velocities of the cone motors. And the rest are the angular velocities of the rotor motors.

III. Thrusts and Drag Moments

This section explains the method we model the thrust and the drag moments.

Several studies (e.g., [20,21]) take the effect of blade flapping into consideration. To simplify our modeling process, this effect is beyond the scope of this research and can be a further topic.

Typical formulas modeling the thrust and drag moment in are in Equation (19)-(20), respectively.

$$F_i = k_f \cdot w_i^2 \quad (19)$$

$$M_i = k_m \cdot w_i^2 \quad (20)$$

These thrust and drag moment models are generally used in the conventional quadrotors [22,23] and the tilt UAV [24]. It is worth mentioning that $w_i$ in Equation (19) and (20) in a conventional quadrotor is the angular velocity of the propeller with respect to the body-fixed frame. Although a more reasonable way is to deduce the angular velocity with respect to the inertial frame, we still deduce it with respect to the body-fixed frame.

The angular velocity of the propeller in the quad-tilt-rotor is a little different. In [9], the drag-moment-related angular velocity is with respect to the body-fixed frame. While the thrust-related angular velocity is with respect to its previous frame; it is the angular velocity of the speed motor.

In this research, both the drag-moment-related and the thrust-related angular velocities of the quad-cone-rotor are with respect to the body-fixed frame.

*Proposition 5:* (thrust and drag moment) The thrust and the drag moment in quad-cone-rotor are modeled in Equation (21)-(22).

$$F_i = k_{f_i} \cdot ({}^{R_i}w_{Z_i}^{B})^2 \quad (21)$$

$$M_i = k_{m_i} \cdot ({}^{R_i}w_{Z_i}^{B})^2 \quad (22)$$

$^{R_i}w^B_{Z_i}$ in Equation (21)-(22) is the part of the angular velocity of the propeller along $z-direction$ with respect to the body-fixed frame, $\mathcal{F}_B$, expressed in the rotor frame, $\mathcal{F}_{R_i}$.

For example, the angular velocity of the propeller with respect to the body-fixed frame, $^{R_i}w^B_i$, in (23) means that there are rotations along $X-axis, Y-axis, Z-axis$. The $^{R_i}w^B_{Z_i}$ required in Equation (21)-(22) is the part of the angular velocity along $Z-axis$ of $^{R_i}w^B_i$ (Equation (24)).

$$^{R_i}w^B_i = \begin{bmatrix} 1 \\ 2 \\ 3 \end{bmatrix} \quad (23)$$

$$^{R_i}w^B_{Z_i} = 3 \quad (24)$$

$^{R_i}w^B_i$ can be calculated similarly with Equation (8). However, the first term in Equation (8) vanishes in $^{R_i}w^B_i$ since $^{R_i}w^B_i$ is with respect to the body-fixed frame. The result is in Equation (25).

$$^{R_i}w^B_i = \left[R_Z\left((-1)^i \cdot \theta_i\right) \cdot R_X(\phi)\right]^T \cdot \omega^B_{C_i} + \begin{bmatrix} 0 \\ 0 \\ (-1)^i \cdot \varpi_i \end{bmatrix}$$

$$= (-1)^i \cdot \begin{bmatrix} 0 \\ \dot{\theta}_i \cdot \sin(\phi) \\ \dot{\theta}_i \cdot \cos(\phi) + \varpi_i \end{bmatrix} \quad (25)$$

Notice that there are rotations along $Y-axis$ and $Z-axis$. *Proposition 5* believes that only the angular velocity along $Z-axis$ contributes to the thrust and drag moment. The term $^{R_i}w^B_{Z_i}$ is in Equation (26).

$$^{R_i}w^B_{Z_i} = (-1)^i \cdot \left(\dot{\theta}_i \cdot \cos(\phi) + \varpi_i\right) \quad (26)$$

It can be seen from Equation (26), not only the angular velocity of the rotor-motor, $\varpi_i$, which directly drives the propeller but also angular velocity of the cone-motor, $\dot{\theta}_i$, which changes the direction of the thrust contribute to $^{R_i}w^B_{Z_i}$. When all four rotor-motors suffer from stuck (complete loss in thrusts. $\varpi_i = 0$, $i = 1, 2, 3, 4$), the components of $\dot{\theta}_i$ is still able to provide thrusts to some extent, making possibility of the fault-tolerant control in this severe case.

Besides the angular velocity, $^{R_i}w^B_i$, another quantity worth investigation is the translational velocity. The translational velocity of the geometric center of the propeller with respect to the body-fixed frame, $\mathcal{F}_B$, expressed in the rotor frame, $\mathcal{F}_{R_i}$, is Equation (27).

$$^{R_i}v^B_{P_i} = (-1)^i \cdot \begin{bmatrix} \sin(\phi) \cdot d \cdot \dot{\theta}_i \\ 0 \\ 0 \end{bmatrix} \quad (27)$$

$d$ in Equation (27) is the length between the origin of $\mathcal{F}_{R_i}$ and the origin of $\mathcal{F}_{C_i}$. We ignore this term in our analysis since it can be ignored if $d$ is designed small.

The angular velocity, $^{R_i}w^B_i$, and the translational velocity, $^{R_i}v^B_{P_i}$, expressed in the rotor-motor frame, $\mathcal{F}_{R_i}$, are exemplified for the second motor rotor frame, $\mathcal{F}_{R_2}$, in Figure 8.

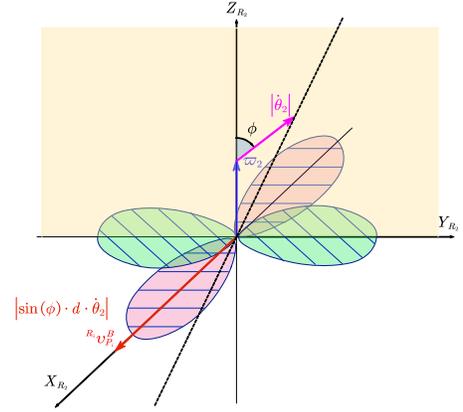

Fig. 8. The angular velocity and the translational velocity in the rotor frame $\mathcal{F}_{R_2}$

IV. Symmetric Open-loop Hovering

The closed-loop controller design is beyond the scope of this research. We only analyze the symmetric open-loop hovering test in this study. We restrict our study in the case with the fixed direction of the cone-motors $\theta_1 \equiv \theta_3 \equiv 0, \theta_2 \equiv \theta_4 \equiv \pi$.

Our goal is to find the relevant angular velocities of the rotor-motors ($\varpi_i$, $i = 1, 2, 3, 4$) for different designed cone-angle, $\phi$.

The parameters used in this simulation are given in Figure 9.

| length of the arm | 0.1785 meter |
| --- | --- |
| d | 0.01 meter |
| g | 9.8 N/kg |
| total mass | 0.429 kg |
| I$_B$ | diag(2.238E-3, 2.985E-3, 4.804E-3) kgm² |
| I$_{C_i}$ | diag(1E-10, 1E-10, 2.030E-5) kgm² |
| I$_{R_i}$ | diag(1E-10, 1E-10, 2.030E-5) kgm² |
| k$_{m_i}$ | 2.423E-7 Nm |
| k$_{f_i}$ | 8.048E-6 N |

Fig. 9. The parameters in the Quad-cone-rotor.

Equation (28)-(29) elucidate relationship between the angular velocities of the rotor-motors, $\varpi_i (i=1,2,3,4)$, in a symmetric open-loop hovering.

$$\varpi_1 = \varpi_3 = \sqrt{\frac{m \cdot g}{4 \cdot k_f}} \qquad (28)$$

$$\varpi_2 = \varpi_4 = \sqrt{\frac{m \cdot g}{4 \cdot k_f \cdot \cos(2\phi)}} \qquad (29)$$

The result is plotted in Figure 10. The cone-angle varies from 0 radius to $\frac{\pi}{4}$ radius (asymptote). The blue curve and the red curve are the angular velocity in Equation (28) and (29), respectively. The red stars and the blue stars are the situations verified in the quad-cone-rotor simulator; the zero vertical acceleration is received for these angular velocities with the corresponding cone-angles.

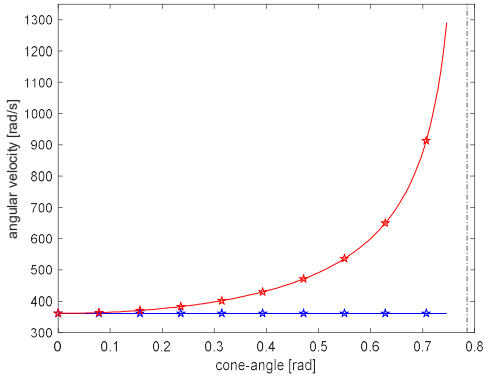

Fig. 10. Angular velocities in symmetric open-loop hovering.

## V. Fault-tolerant Open-loop Hovering

Fault diagnosis and fault-tolerant control received significant interests in quadrotor and tilt quadrotor control. [25] analyzes the case where two rotors suffer from complete thrusts loss and provides fault-tolerant control methods with a free yaw in a conventional UAV. To solve the uncontrolled yaw in [25], [11] proposed a tilt structure providing the possibility of yaw control.

However, there lacks the study for the case where all the thrusts suffer from complete loss. The reason is that controlling a conventional quadrotor or a quad-cone-rotor suffering from such a severe fault is impossible. We need the admissible thrusts [26,27] to realize the attitude and the altitude control; zero thrust in all rotors is not within the admissible region.

However, the situation is different in our quad-cone-rotor. It can be seen from Equation (25)-(26), even if all the rotor-motors suffer from complete thrust loss ($\varpi_i = 0, i=1,2,3,4$), the actual thrust can be non-zero if we assign the $\dot{\theta}_i$ properly with a non-zero cone-angle ($\phi \neq 0$). This indicates that a fault-tolerant control is possible in a quad-cone-rotor while all the rotors suffer from a total loss in thrust.

In this study, we analyze the fault-tolerant ability in fulfilling an open-loop hovering when all the speed-motors are unhealthy (complete loss in thrust, $\varpi_i = 0, i=1,2,3,4$). The quad-cone-rotor generates the thrusts totally relying on the angular velocities of the cone-motors ($\dot{\theta}_i, i=1,2,3,4$).

Likewise, the detail of closed-loop fault-tolerant control is one of the further studies and is not considered in this research.

*Proposition 6:* (cone motor and speed motor) Cone motor has two modes. One is working as a servomotor; the other mode is the speed motor. In fault-tolerant control, the cone motors are working as speed motors. The failed ($\varpi_i = 0$) rotor motors are fixed at the positions generating the largest thrust.

Equation (30)-(31) illustrate the relationships of the angular velocities of the cone motors and the rotor motors, respectively, a complete-loss-in-all-thrust fault.

$$\dot{\theta}_1 = \dot{\theta}_2 = \dot{\theta}_3 = \dot{\theta}_4 = \dot{\theta}_c (constant) \qquad (30)$$

$$\varpi_i = 0, i=1,2,3,4 \qquad (31)$$

*Proposition 7:* (cone motor and speed motor) The angular velocity of the cone motor, $\dot{\theta}_c$, results from Equation (32)-(33).

$$A(t) = \int \left[ k_f \cdot \left(\dot{\theta}_c \cdot \cos(\phi)\right)^2 \cdot \left(1 - 2 \cdot \sin^2(\phi) \cdot \sin^2\left(\frac{\dot{\theta}_c \cdot t}{2}\right)\right) - \frac{1}{4} mg \right] \cdot dt \qquad (32)$$

$$\int_0^{\frac{2\pi}{\dot{\theta}_c}} A(t) \cdot dt = 0 \qquad (33)$$

Proof: see Appendix

Instead of solving Equation (32)-(33), we find $\dot{\theta}_c$ in the quad-cone-rotor simulator to make it near hovering for different cone-angle, $\phi$. The result is plotted in Figure 11.

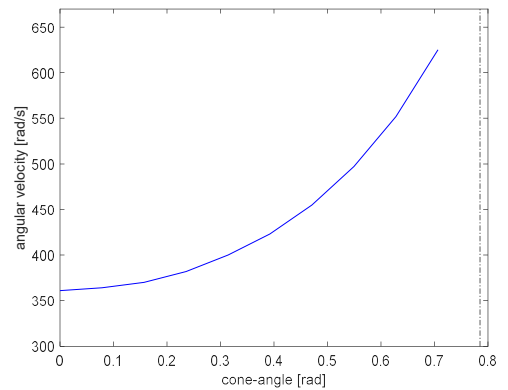

Fig. 11. Angular velocities in fault-tolerant open-loop hovering

It is worth mentioning that an intensive oscillation is observed in the vertical acceleration for any cone-angle, $\phi$. Figure 12a is the vertical acceleration for the case $\phi = \frac{\pi}{10}$. It is reasonable since that the $Z-axis$ projection of the thrust varies, inducing the changing vertical acceleration, (Equation (32)-(33)). Consequently, the oscillation in the vertical acceleration occurs.

The frequency in that oscillation is identical to the frequency of the angular velocity of the cone motor, $\dot{\theta}_c$. This is verified by the power spectral density analysis (Figure 12b and Figure 13).

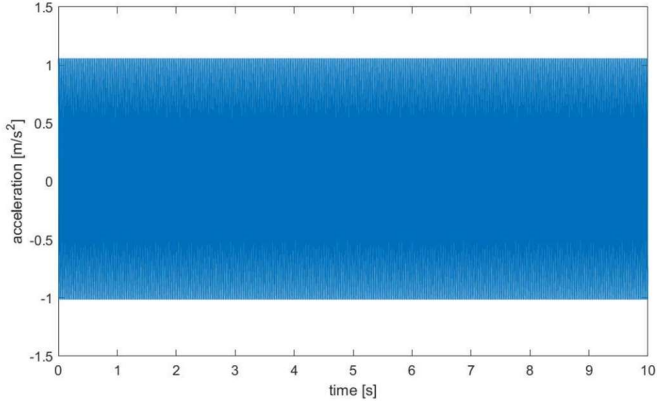

(a)

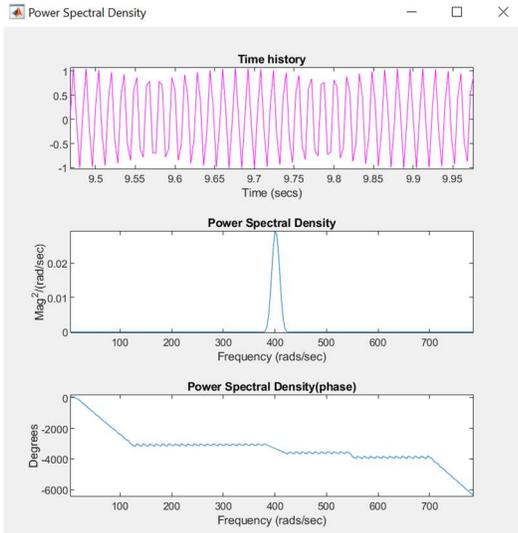

(b)

Fig. 12. The oscillation in acceleration ($\phi = \frac{\pi}{10}$)

We calculate the square of the magnitude (power) of the coupled oscillation for each corresponding cone angle, $\phi$, in each fault-tolerant open-loop hovering test. The $y-axis$ in Figure 13 represents the power. The $X-axis$ represents the angular velocity for the corresponding cone angle.

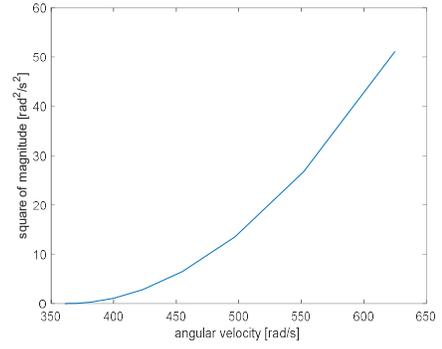

Fig. 13. The power of the coupled oscillation in different angular velocities.

## VI. Cone-angle Trade-off

The cone-angle in the quad-cone-rotor, $\phi$, is one of the most important parameters. It determines the range that the direction of the thrust can be selected. A naïve thought is to design it as large as possible. However, a $\phi$ larger than $\frac{\pi}{4}$ $rad$ generates no positive vertical thrust when for some angles in the cone-motor. In this consideration, we restrict $\phi$ to be no more than $\frac{\pi}{4}$ $rad$.

Another fact that should pay attention to is the centripetal force. Before calculating the centripetal force of a motor while working in a fault-tolerant control mode mentioned in the Section VI (Equation (30)-(31)), we make some reasonable approximation to the direction of the thrust to simplify our trade-off analysis.

Assume that the thrust is in the direction pointing from the apex of the cone to the center of the circle of the cone. Notice that this is the approximation only in this trade-off analysis.

Then, the resulting centripetal force, $F_C(\phi)$, for fault-tolerate control (hovering) based on this approximation is Equation (34).

$$F_C(\phi) = \frac{m^2 \cdot g \cdot d}{4 \cdot k_f} \cdot \frac{\sin(\phi)}{\cos^3(\phi)} \tag{34}$$

$F_C(\phi)$ is a monotonically increasing function within the interval $\left[0, \frac{\pi}{4}\right]$, this indicates that the centripetal force increases while picking a larger cone-angle.

Another parameter affects the centripetal force is $d$. It is the length between the origin of $\mathcal{F}_{R_i}$ and the origin of $\mathcal{F}_{C_i}$. A smaller $d$ not only produces less $F_C(\phi)$ in Equation (34) but also gives less translational velocity in (27). Thus, a small $d$ is encouraged.

To balance the conflict in larger range in direction and the less centripetal force, we formulate the following multiple-objective optimization problem in Equation (35).

$$\begin{cases} Maximize: \text{ range } R(\phi) = 2\pi \cdot d \cdot \sin(\phi) \\ Minimize: F_C(\phi) = \frac{m^2 \cdot g \cdot d}{4 \cdot k_f} \cdot \frac{\sin(\phi)}{\cos^3(\phi)} \\ Subject\ to: 0 \leqslant \phi \leqslant \frac{\pi}{4} \end{cases} \tag{35}$$

The way we solve this multiple-objective optimization problem is the weighted sum method [28], transferring Equation (35) into a single-objective optimization problem in Equation (36).

$$\begin{cases} Minimize: C(\phi) = -R(\phi) + \mu \cdot F_C(\phi), \ (\mu \geq 0) \\ R(\phi) = 2\pi \cdot d \cdot \sin(\phi) \\ F_C(\phi) = \dfrac{m^2 \cdot g \cdot d}{4 \cdot k_f} \cdot \dfrac{\sin(\phi)}{\cos^3(\phi)} \\ Subject\ to: \ 0 \leq \phi \leq \dfrac{\pi}{4} \end{cases} \quad (36)$$

The result is plotted in Figure 14. It is a Pareto frontier of the trade-off between the range and the centripetal force. Since the $X-axis$ represents the negative of the range, the less (more negative) the better. The $Y-axis$ represents the centripetal force, the less the better. However, decreasing the centripetal force increases the negative range, causing the range loss for the direction of the designed thrust.

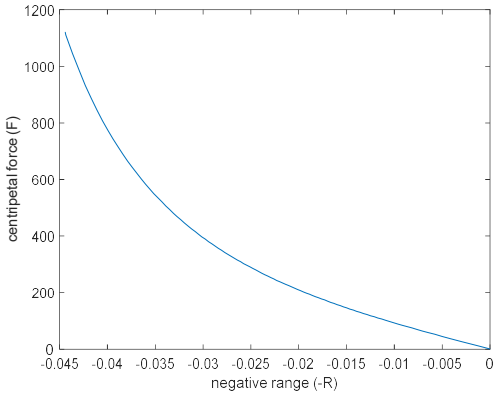

Fig. 14. The Pareto frontier of the trade-off between the range and the centripetal force.

VII. Conclusion and Discussion

The quad-cone-rotor realizes the open-loop hovering in a healthy state. Moreover, it shows unique potential in fault-tolerant control when a severe fault happens (complete loss in all thrusts). The oscillation is observed while taking this special fault-tolerant control. The frequency of this oscillation is identical to the frequency of the angular velocity.

Our ongoing further works are closed-loop control and fault-tolerant control. It might be tempting to control the translational movement by setting $\theta_1 \equiv \theta_2 \equiv \pi, \theta_3 \equiv \theta_4 \equiv 0$. In this setting, the quad-cone-rotor can balance the altitude while receiving a level force, making it move like an autonomous car. However, the drag moment cannot be balanced, making it rotate.

Another special restriction in quad-cone-rotor is the admissible analyzing control input. Judged from Equation (25)-(26), it seems that $\dot{\theta}_i \cdot \cos(\phi)$ and $\varpi_i$ have the same effect affecting the angular velocity generating the thrust. However, pairing $\dot{\theta}_i$ and $\varpi_i$ in some relationships are prohibited. When $\varpi_i$ is not very large, the rotating propeller can be a burden to generate a model of the thrust; the changing direction of the propeller generates different thrust in each posture. That is also part of the reason why we make *Proposition 6*.

Although some pairs generate thrust, modeling it using a precise mathematical way can be challenged. To better design our controller, we make the following restriction to our inputs in Equation (37).

$$\begin{cases} \varpi_i \geq 10 \cdot |\dot{\theta}_i|, \quad normal\ working\ mode \\ \varpi_i \equiv 0, \dot{\theta}_i\ unrestricted, \quad loss\ in\ thrust \end{cases} \quad (37)$$

When the quad-cone-rotor is working in the normal mode, the $\varpi_i$ is much larger than $\dot{\theta}_i$. This restriction assigns $\varpi_i$ to be the angular velocity mainly for generating the magnitude of the thrust. While $\dot{\theta}_i$ serves as a direction adjuster for the thrust. The other admissible working mode is usually applicable for a fault-tolerant control. The only angular velocity allowed is $\dot{\theta}_i$. While $\varpi_i$ is prohibited for introducing the difficulty in analysis.

The relevant admissible region is painted in blue in Figure 15. It consists of the triangular part in the first and second quadrants and the positive direction of the $X-axis$.

Our further research will base on this admissible region to design the relevant controllers.

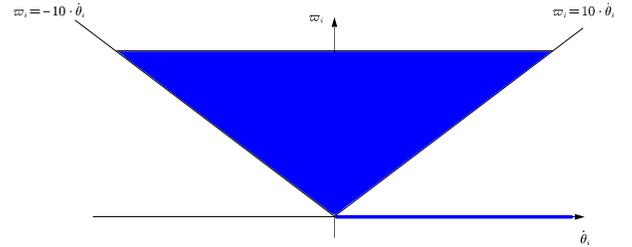

Fig. 15. The admissible input region of the quad-cone-rotor.

**Appendix**: Proof of *Proposition 7*

To prove *Proposition 7*, we first prove the following *lemma*.

*Lemma:* The included angle ($\kappa$) between the direction of the thrust and the $Y-axis$ in the rotor frame is Equation (38).

$$\kappa = \arccos\left(1 - 2 \cdot \sin^2(\phi) \cdot \sin^2\left(\dfrac{\theta}{2}\right)\right) \quad (38)$$

*Proof of Lemma:*

The relationship of each quantity is plotted in Figure 16.

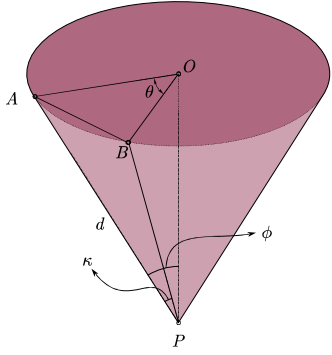

Fig. 16. The relationship between $\theta$, $\phi$, $\kappa$.

We calculate $OA$ and $OB$ in Equation (39) based on the geometric relationship in a cone.

$$OA = OB = d \cdot \sin(\phi) \tag{39}$$

In *isosceles* $\triangle OAB$, we have Equation (40).

$$AB = 2 \cdot OA \cdot \sin\left(\frac{\theta}{2}\right) \tag{40}$$

In *isosceles* $\triangle PAB$, we have Equation (41), law of cosines.

$$\cos(\kappa) = \frac{PA^2 + PB^2 - AB^2}{2 \cdot PA \cdot PB} \tag{41}$$

Substitute Equation (39)-(40) into Equation (41), we receive Equation (38). The *proof to lemma* is complete.

Each rotor in the symmetric hovering analysis satisfies Equation (42), Newtown's second law.

$$\frac{1}{4} \cdot m \cdot \ddot{Z} = T_i \cdot \cos(\kappa) - \frac{1}{4} \cdot m \cdot g \tag{42}$$

Substituting Equation (21), (26), (38) into Equation (42) yields Equation (43).

$$\frac{1}{4} \cdot m \cdot \ddot{Z} = k_f \cdot \left(\dot{\theta}_c \cdot \cos(\phi)\right)^2 \cdot \left(1 - 2 \cdot \sin^2(\phi) \cdot \sin^2\left(\frac{\dot{\theta}_c \cdot t}{2}\right)\right) - \frac{1}{4} mg \tag{43}$$

Integrating the right side of Equation (43) yields the velocity-related term, Equation (32).

Integrating Equation (32) again yields the altitude-related term, Equation (33). The expected altitude change in a period, $\frac{2\pi}{\dot{\theta}_c}$, is zero.

So far, the *Proposition 7* is proved.